\documentclass[runningheads]{llncs}
\usepackage{cite}
\usepackage{soul}
\usepackage{hyperref}
\usepackage{multirow}
\usepackage{booktabs}
\usepackage{amsmath,amssymb,amsfonts}
\usepackage{algorithmic}
\usepackage{graphicx}
\usepackage{textcomp}
\usepackage{makecell}
\usepackage{xcolor}
\usepackage[T1]{fontenc}
\usepackage{graphicx}
\begin{document}
\title{Drugs Resistance Analysis from Scarce Health Records via Multi-task Graph Representation}


\author{Honglin Shu\inst{1,*} \and 
Pei Gao\inst{2,3,*} \and
Lingwei Zhu\inst{4} \and
Zheng Chen\inst{5} }
%
\authorrunning{F. Author et al.}
%
\institute{Hong Kong Polytechnic University, Hong Kong SAR, China \and
Sony AI, Tokyo, Japan \and
Nara Institute of Science and Technology, Ikoma, Japan \and
University of Alberta, Edmonton, Canada \and
Osaka University, ISIR, Osaka, Japan\\
\email{*Corresponding: honglin.shu@connect.polyu.hk, \\ gao.pei.gi3@is.naist.jp}
}
\maketitle              
\begin{abstract}
Clinicians prescribe antibiotics by looking at the patient's health record with an experienced eye. However, the therapy might be rendered futile if the patient has drug resistance. Determining drug resistance requires time-consuming laboratory-level testing while applying clinicians' heuristics in an automated way is difficult due to the categorical or binary medical events that constitute health records. In this paper, we propose a novel framework for rapid clinical intervention by viewing health records as graphs whose nodes are mapped from medical events and edges as correspondence between events in given a time window. A novel graph-based model is then proposed to extract informative features and yield automated drug resistance analysis from those high-dimensional and scarce graphs. The proposed method integrates multi-task learning into a common feature extracting graph encoder for simultaneous analyses of multiple drugs as well as stabilizing learning. On a massive dataset comprising over 110,000 patients with urinary tract infections, we verify the proposed method is capable of attaining superior performance on the drug resistance prediction problem. Furthermore, automated drug recommendations resemblant to laboratory-level testing can also be made based on the model resistance analysis. 

\end{abstract}
\section{Introduction}

\begin{figure}[t]
\centering
\includegraphics[width=0.92\linewidth]{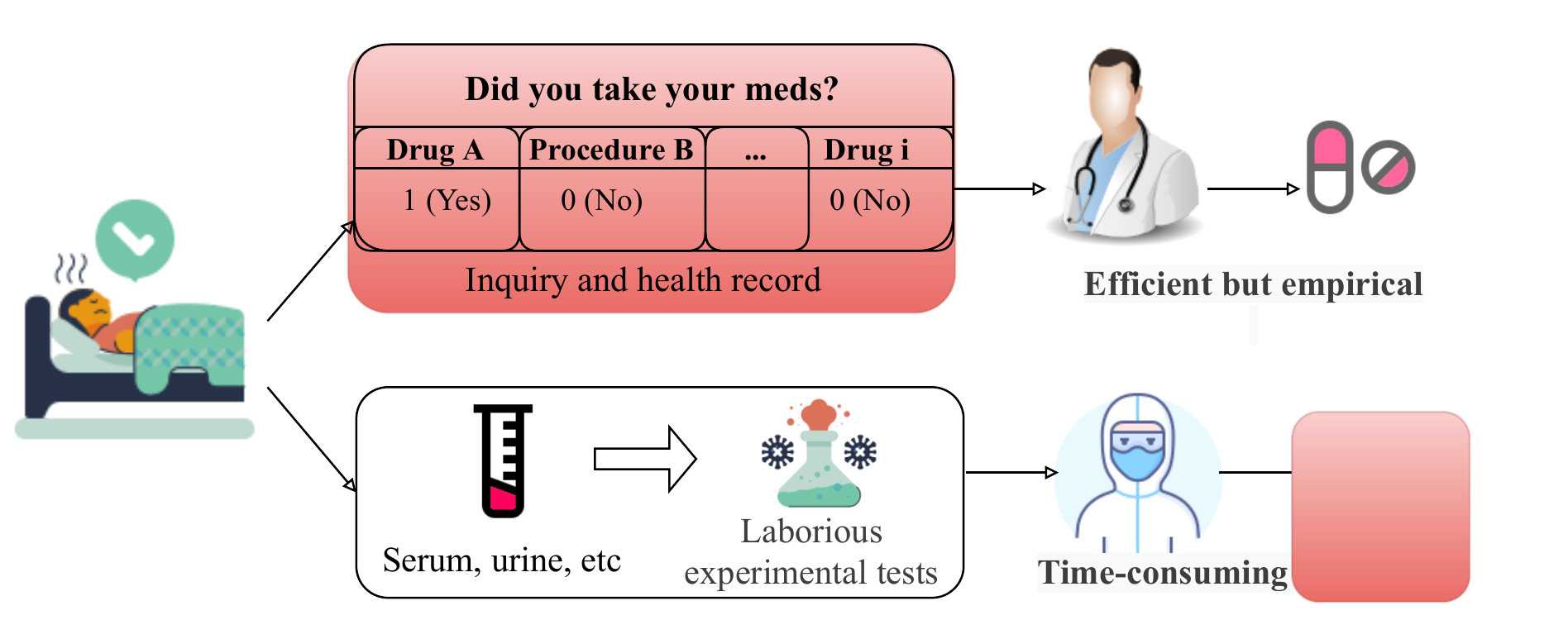}
\caption{Clinicians prescribe antibiotics by checking the patient's health record consisting of binary medical events. 
Drug resistance is determined in microbiology labs after time-consuming testing.
Hence, the purpose of this work is to investigates \emph{\underline{"how to provide a precise drug recommend by binary health record data?"}}}
\label{fig:story}
\end{figure}

Multi-drug resistance, i.e., the insusceptibility to multiple antimicrobial agents in pathogenic bacteria, is currently endangering the efficacy of antibiotic treatment and has become a staple global concern in public health \cite{ventola2015antibiotic}.
Antibiotics are usually chosen empirically and their resistance needs to be determined via laboratory-level drug resistance testing using specimens such as serum, urine, cerebrospinal fluid, etc. as shown in Fig. \ref{fig:story}. \cite{reller2009antimicrobial}.
While laboratory testing has shown its indisputable efficacy in clinical practice, rapid clinical intervention such as drug recommendation becomes impossible due to the long hours in microbiology laboratories that can last up to 72 hours.
Further, it is often difficult to apply in large scale the heuristics of human clinicians \cite{ferrer2014empiric}.

The community has seen a large number of research attempts on developing data-driven clinical assistants to accurately identify candidates for the therapy. 
Among these works, there is a trend on combining machine learning (ML) methods with Health Record (HR) data for more accurate and adaptive prediction of drug resistance \cite{yelin2019personal}. 
Although there is promise on the reported performance, those traditional statistical ML models have not yet been widely adopted in real clinic scenes.
In general, statistical ML models generate probabilistic predictions with a focus on minimizing the number of misclassifications. 
But what is more relevant for causal inference in clinical practice lies in understanding the data-representation process. 
A good representation makes it easier to extract crucial information when building classifiers or any kind of predictors, 
as well as to contribute to better model performance, interpretability and credibility.

The HR data can be represented as a series of patient status information (medical events) with a time dimension (time windows).
These medical events are usually recorded as a row in a table with each element being "binary" to denote whether a medical event \emph{happens or not}. 
Such $\{0,1\}$ encoded tabular data, though intuitive, incurs intractable computational burden e.g., when using dictionaries for mapping categorical features to vector indices \cite{bertsimas2020sparse}.
In general, high dimensional binary features can cause computability concerns for conventional machine learning models \cite{attenberg2009collaborative}.
Although there are some attempts on solving this issue like feature hashing \cite{seger2018investigation}, they run at the cost of losing latent correlation between medical event features due to  irreversibility and unlinkability \cite{tulyakov2007symmetric}.

The troublesome inexpressive binary data turn out to have surprisingly suitable representations as graphs: since each row of the HR is a series of binary elements showing a patient's historical medical events, we can consider it as a graph with events being nodes and edges reflecting binary relationship between two nodes, i.e., an edge exists when two medical events happened at the same time.
Graph Neural Networks (GNNs) \cite{Wu2019ACS} have recently made significant progress in processing graph data and finding hidden information. 
The majority of GNNs \cite{DBLP:conf/iclr/KipfW17, DBLP:conf/icml/WuSZFYW19, DBLP:conf/nips/HamiltonYL17,DBLP:conf/iclr/VelickovicCCRLB18,DBLP:conf/iclr/XuHLJ19} use a message passing mechanism \cite{DBLP:conf/icml/GilmerSRVD17}. 
The Graph Isomorphism Network (GIN) \cite{DBLP:conf/iclr/XuHLJ19}, for example, uses an aggregation function to iteratively update the node representations before generating the vector representation of the graph via the readout function \cite{DBLP:conf/nips/YingY0RHL18, DBLP:conf/kdd/0001WAT19}. 
GNN can effectively learn from graph structure information to more expressive latent vector representations using this form to improve the performance of downstream tasks. 
The medical event graph can help to learn a more telling latent representation by providing additional structure information.

However, applying GNNs to health records is far from being straightforward.
Another majoy difficulty arises from the fact that the emergence of drug resistance is a low probability event, hence supervisory signals one can exploit to train the model is highly imbalanced, with far more negative samples than positive ones. 
As a result, noise in the supervised signals has a significant impact on supervised learning, causing biased learning and overfitting to the partial label distribution.
To solve this problem, we take inspiration from the recently popular multi-task learning (MTL) literature. 
MTL with neural networks is accomplished by learning shared representations from various supervisory signals of many tasks to reduce the impact of noise on any single task \cite{DBLP:journals/pami/VandenhendeGGPD22}.

In this paper we propose a novel framework for handling massive health records (more than 110,000 patients) from which we extract useful information such as logical correspondence between highly sparse medical events.
By viewing each patient as a giant graph with binary medical events (and their connections) being nodes (and edges), our framework is capable of leveraging graph properties to extract informative features which conventional methods struggle to do.
We further propose to alleviate the impact of imbalanced supervisory signals resulted from the sparsity of the dataset by multi-task learning.
Our novel architecture consists of a shared input phase and a multi-head output that yields prediction for each of the \emph{laboratory-recommended antibiotics}.
In summary, the contributions of this paper are:
\begin{itemize}
    \item We propose a novel graph learning framework that extracts informative features from high-dimensional sparse health records by transforming patient status to graphs. 
    The multi-head architecture based on MTL outputs predictions for all labels exploiting one model and alleviates great impact imposed by imbalanced supervisory signals.
    \item On a massive dataset consisting of more than 110,000 patients we verify the proposed framework is capable of learning informative correspondence between medical events and outputting antibiotics recommendations highly similar to laboratory labels.
\end{itemize}


\section{Graph Neural Networks and Multi-Tasks Learning}

\textbf{Graph Neural Networks (GNNs)} leverage the graph structure information and node attributes to generate the latent representation of each node \(\mathbf{h}_{v}\) or the entire graph \(\mathbf{h}_{\mathcal{G}}\). A message passing mechanism is used by the majority of modern GNNs to iteratively aggregate the representation of each node's neighbors and update the latent representation of each node \cite{DBLP:conf/icml/GilmerSRVD17}. The latent representation \(\mathbf{h}_{v}^{k}\) of node \(v\) after \(k\) iterations of aggregation is summarized as:
\begin{align}
  \mathbf{h}_{v}^{k} = \Delta ^{(k)}\left(\mathbf{h}_{v}^{k-1}, \mathbf{a}_{v}^{k}\right),
   \text{where       }  \mathbf{a}_{v}^{k} = \delta^{(k)}\left({\mathbf{h}_{u}^{k-1}\, |\, {u} \in \mathcal{N}(v)}\right).
\end{align}

\noindent
where $\Delta$ is the update operation, 
$\delta$ is the aggregation.
\(\mathcal{N}(v)\) is a set of neighbors of node \(v\) and \(\mathbf{h}_{v}^{0}\) is initialized as node attributes or one-one encoding when node attributes is not available. Latent representation \(\mathbf{h}_{v}^{k}\) is used to do node-level tasks such as node classification or link prediction. 
For graph classification, we need to combine all node representations in graph \(\mathcal{G}\) to generate a representation \({h}_{\mathcal{G}}\) of the entire graph via the readout function.
The expressivity of latent representations depends on the choice of the aggregation, update, and readout functions. Many different aggregation function have been suggested. 
For instance, in Graph Convolutional Networks (GCN) \cite{DBLP:conf/iclr/KipfW17},  the average neighbor messages aggregation is employed to learn node representation.
Simplifying Graph Convolution (SGC) \cite{DBLP:conf/icml/WuSZFYW19} eliminates superfluous complexity by progressively reducing nonlinearities and weight matrices between succeeding layers. 


\noindent\textbf{Multi-task learning (MTL)} refers to more than one loss function being optimized during model training.
MTL can improve the model's generalizability on all tasks by allowing the model to share the domain-specific representations it has learned during training on a series related tasks \cite{DBLP:journals/pami/VandenhendeGGPD22}. 
The key component is to design an efficient parameter-sharing strategy, including a share on top of an encoder with numerous task-specific decoders \cite{DBLP:conf/nips/SenerK18}, namely hard-parameter sharing. 
Another typical strategy is the soft-parameter sharing, here, each task has its own independent parameters in the representation extraction layers and the decision layers, they express similarity by restricting the discrepancies between the parameters of various tasks. 
Some recent works are developed with tensor trace norms to discover the low-rank structure \cite{DBLP:conf/kdd/ZhangZW21}, or attention mechanism to share feature \cite{DBLP:conf/icdm/PhamYMZZ21}. Inspired above description, we consider MLT suitable for the analysis of Multi-drug resistance, and we position our research problem and detail the proposed method as follows.

 



\section{Problem Formulation}
\label{sec:problem setting}
Our objective is to investigate how to accurately evaluate whether a patient has resistance to a variety of medications based on their medical history. Let matrix \(\mathbf{X} \in \mathbb{R}^{M \times N}\) be a dataset and \(\mathbf{y} = \{\mathbf{y}_{n} \in \mathbb{R}^{M} : n \in \{1, 2,..., T\}\}\) denotes the set of labels for the patients' drug resistance, where $M$ is the number of patients and $N$ medical events. 
The elements of matrix \(\mathbf{X}\) are binary, i.e., if a specific medical event \(i\) occurred in certain samples \(j\), the corresponding element \(\mathbf{X}_{i,j}\) is 1, otherwise it is 0. The \(i\)-th element \(\mathbf{y}_{n}^{(i)}\) of \(\mathbf{y}_{n}\) is 1 if patient \(i\) has resistance to drug \(n\), otherwise it is 0. 
\(T\) represents the number of distinct drug resistances that must be predicted (i.e., number of tasks).

The matrix \(\mathbf{X}\) is significantly sparse and high-dimensional. Conventional methods often struggle to effectively estimate the conditional probability distribution \(P\left(\mathbf{y}_{n} \,\big|\,\mathbf{X},\,{\theta}\right)\), where $\theta$ denotes the parametrization of the model.
To alleviate the issue of sparsity and high-dimensionality, we convert each row of \(\mathbf{X}\) into a medical event graph \(\mathcal{G}_{i}\) with \(i \in \{1,2,...,M\}\), where the nodes in \(\mathcal{G}_{i}\) denote the historical medical events and the connectivity of two nodes depends on whether these two medical events happened at the same time. 
We use a graph-based encoder \(f_{\mathcal{G}, {\theta}}\left(\cdot\right)\) to learn an expressive graph vector representation since one can mine more information from its topological structure than digging from raw data. 
Therefore, the problem is modeled as a graph classification task equivalent to a graph isomorphism problem:
\begin{align}
  {\theta}_{n} = \arg\max_{\theta} \sum_{i=1}^{M} P\left(\mathbf{y}_{n}^{(i)} \,\big|\, f_{\mathcal{G}, {\theta}}\left(\mathcal{G}_{i}\right)\right), \,\, n \in \{1, 2,..., T\},
\end{align}
i.e., the solution to $n$-th prediction problem is obtained as the maximizer of the conditional probability over all training instances.

More importantly, since the events of drug resistance occur infrequently, the resistance labels of patients to all drugs \(\mathbf{y}\) is unbalanced: \emph{the number of positive samples is substantially smaller than the number of negative samples.}
This type of supervision signals can mislead the model into biased latent representations and subsequently overfitting to the imbalanced label distribution. 
In real-world scenarios, such high precision low recall models is not informative for drug resistance analyses. 

Naturally, the medical decision for assessment of drug resistance is to find a trade-off between multiple drugs. 
Hence, to alleviate the negative effect from the biased supervisory signals of any single task, we can formulate the problem as a multi-objective optimization problem with graph-based shared parameter encoder, which can help the model filter out noises in single-task labels and force the model to extract more generalizable graph latent representations. 
Our proposed objective formulation is to minimize a weighted summation of losses on the prediction of different drugs resistance:
\begin{equation}
  \begin{aligned}
    &{\arg\min}_{\theta_{\mathcal{G}}, \theta_{1}, \theta_{2},...,\theta_{T}} \mathcal{L}_{Total}\left(\theta_{\mathcal{G}}, \theta_{1}, \theta_{2},...,\theta_{T}\right)\\
    &= {\arg\min}_{\theta_{\mathcal{G}}, \theta_{1}, \theta_{2},...,\theta_{T}} \sum_{n=1}^{T} {w}_{n} \mathcal{L}_{n}\left(\theta_{\mathcal{G}}, \theta_{n}\right),
  \end{aligned}
\end{equation}
where \(\theta_{\mathcal{G}}\) denote the shared parameters of graph-based encoder and \(\theta_{n}\) the independent parameters of task-specific learners. 
\({w}_{n}\) is a weight coefficient of each loss to control the difficulty and speed in model learning. 

\begin{figure}[t]
\centering
\includegraphics[width=1.0\linewidth]{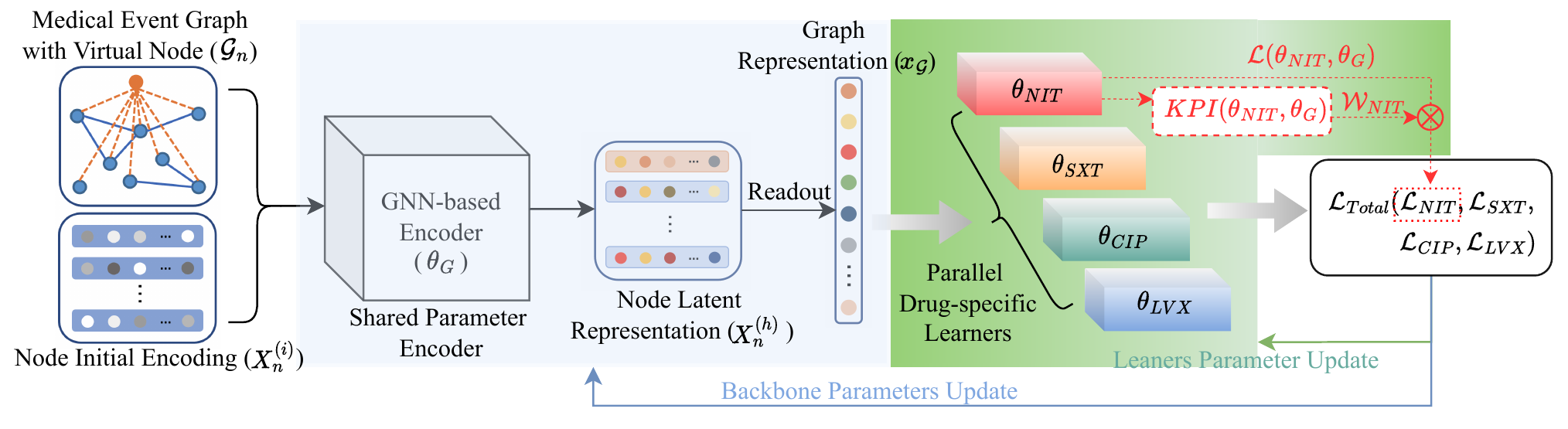}
\caption{System overview of the proposed GNN-based dynamic weighted multi-task learning framework.}
\label{fig:systemOV}
\end{figure}

\section{Proposed Framework}
\label{sec:proposal}
As illustrated in Fig. \ref{fig:systemOV}, we propose a novel and effective multi-task graph learning framework leveraging a GNN-based encoder as the bottom parameter-sharing module, parallel task-specific learners as drug-specific resistance predictors, and dynamically weighted adaption as a mechanism to balance learning speed and problem difficulty.

\noindent\textbf{GNN-based Encoder.} Given a vector \(x_i \in \mathbb{R}^{N}\) denoting $N$ historical medical events of the $i$-th patient, we convert it to a medical event graph \(\mathcal{G}_{i}\). 
Let us recall that there are $N$ nodes in graph \(\mathcal{G}_{i}\) to represent the $N$ medical events. 
An edge between two nodes exists only when the two corresponding medical events occur at the same time. 

Since the vector \(x_i\) may be very sparse, the corresponding graph \(\mathcal{G}_{i}\) is far from fully connected, and can lead to very poor graph representations if conventional massage-passing-based GNNs are naively used.

To takcle this issue, we add to the graph virtual nodes connected to all other nodes. 
The virtual nodes allow isolated connected components to indirectly update the node representation through it. 
We adopt the GIN convolution \cite{DBLP:conf/iclr/XuHLJ19} as the GNN-based encoder.
The nodes in graph \(\mathcal{G}_{i}\) are using one-hot encoding only, which implies only the graph structure information is exploited to classify a graph. 
GIN imitates the Weisfeiler-Lehman test \cite{Shervashidze2011WeisfeilerLehmanGK} that upper-bounds the graph classification problem with only structural information.
Hence, the $k$-th layers of GNN-based encoder can be expressed as:
\begin{equation}
  \begin{aligned}
    &{h}_{v}^{(k)} = \text{MLP}^{(k)} \left({{h}_{v}^{(k-1)} + \sum_{u \in \mathcal{N}(v)} {h}_{u}^{(k-1)}}\right),
  \end{aligned}
\end{equation}
where \({h}_{v}^{(k)}\) is the vector representation of node \(v \in \mathcal{G}_{i}\) and \(\mathcal{N}(v)\) is the set of nodes adjacent to node \(v\). 
\(\text{MLP}^{(k)}\) is an injection function consisting of $n$ fully connected layers. 
After \(K\) iterations of updating, we can separately readout the node latent representation of each layer via mean pooling to finally generate the a set of graph representations:
\begin{equation}
  \begin{aligned}
    {h}_{\mathcal{G}_{i}} = \{{h}_{\mathcal{G}_{i}}^{(1)}, {h}_{\mathcal{G}_{i}}^{(2)},..., {h}_{\mathcal{G}_{i}}^{(K)}\} 
    = \{\text{MEAN}(\{{h}_{v}^{k} \,|v\, \in \mathcal{G}_{i}\}) \,\big|\, \forall k\}.
  \end{aligned}
\end{equation}

\noindent\textbf{Task-specific Learners.} 
Our model consists of independent tasks-specific learners. 
After the encoder, we simply append linear layers as task-specific learners for predicting drug resistance. 
Given a set of graph representations \({h}_{\mathcal{G}_{i}}\), the probability of resistance of drug \(t\) is calculated as:
\begin{equation}
  \begin{aligned}
    &P\left(\mathbf{y}_{t}^{(i)} = 1 \,\big|\, f_{\theta_{\mathcal{G}}}\left(\mathcal{G}_{i}\right)\right) = \frac{1}{1+e^{-\sum_{k=1}^K {\mathbf{X}_{t} {h}_{\mathcal{G}_{i}}^{(k)}}}},
  \end{aligned}
\end{equation}

\noindent
where \(\mathbf{X}_{t}\) are the task-specific weighted matrix for drug \(t\).

\noindent\textbf{Dynamically Weighted Adaptation.} In multi-task learning, the weights in loss summation will affect task learning difficulty. 
We find that the difficulty of assessing resistance to different drugs also varies. 
Taking inspiration from the focal loss function \cite{DBLP:conf/iccv/LinGGHD17} for sample-level imbalance problem, 
we introduce Dynamic Task Prioritization (DTP) \cite{DBLP:conf/eccv/GuoHHYF18} into the weight adjustment while training our parallel task architecture. DTP adjusts the weight of each task iteratively based on the key performance indicators (KPIs) \(\hat{K}_{t}^{\tau}\). The KPIs \(\hat{K}_{t}^{\tau}\) is computed via moving average:
\begin{equation}
  \begin{aligned}
    \hat{K}_{t}^{\tau} = \alpha {K}_{t}^{\tau} + (1 - \alpha)\hat{K}_{t}^{\tau-1},
  \end{aligned}
\end{equation}

\noindent
where \({K}_{t}^{\tau} \in [0, 1]\) is the performance indicator of task \(t\) on step \(\tau\), e.g., recall. 
\(\alpha\) is a decaying factor. 
Since the number of positive samples is significantly smaller than the amount of negative samples, \emph{recall} has emerged as a key indicator of the learned model's quality  as well as the difficulty of the task to be learnt. 
More specifically, after calculating \(\hat{K}_{t}^{\tau}\), the weight \(w_t^\tau\) for task \(t\) is updated as:
\begin{equation}
  \begin{aligned}
    &{w}_{t}^{\tau} = -(1 - \hat{K}_{t}^{\tau})^\gamma \text{log}\, \hat{K}_{t}^{\tau},
  \end{aligned}
\end{equation}

\noindent
where \(\gamma\) is a hyper-parameter to adjust the weight. 
Higher weights \({w}_{t}^{\tau}\) reflect higher learning difficulty on task \(t\).
The total loss in our proposed MTL framework can hence be summarized as:
\begin{equation}
    \mathcal{L}_{Total} = \sum_{t=1}^{T} {-(1 - \hat{K}_{t}^{\tau})^\gamma \text{log}\, \hat{K}_{t}^{\tau}}\, \mathcal{L}_{t}\left(\theta_{\mathcal{G}}, \theta_{t}\right).
\end{equation}

\noindent\textbf{Optimization.} The weighted sum of loss functions for different tasks has certain favorable properties for multi-task optimization. 
First, it is differentiable, which allows for direct optimization using stochastic gradient descent. 
Second, the chain rule ensures updates of the task-specific learners do not interfere with each other during the optimization process (See Eq. (\ref{eq:loss_each})), and the loss of tasks can update the GNN-based encoders (Eq. (\ref{eq:loss_all})).
\begin{align}
    &\frac{\partial}{\partial \theta_{t}} \mathcal{L}_{Total} = {-(1 - \hat{K}_{t}^{\tau})^\gamma \text{log}\, \hat{K}_{t}^{\tau}} {\frac{\partial}{\partial \theta_{t}} \mathcal{L}_{t}\left(\theta_{\mathcal{G}}, \theta_{t}\right)} \label{eq:loss_each}\\
    &\frac{\partial}{\partial \theta_{\mathcal{G}}} \mathcal{L}_{Total} = \sum_{t=1}^{T} {-(1 - \hat{K}_{t}^{\tau})^\gamma \text{log}\, \hat{K}_{t}^{\tau}} {\frac{\partial}{\partial \theta_{\mathcal{G}}} \mathcal{L}_{t}\left(\theta_{\mathcal{G}}, \theta_{t}\right)} \label{eq:loss_all}
\end{align}
Here, the encoders (green part in Fig. \ref{fig:systemOV}) share the parameters at the same time.

\section{Experiments and Discussion}

\subsection{Dataset Description}
We validate the proposed method on a large-scale AMR-UTI dataset \cite{Michael2020AMR-UTI}.
AMR-UTI dataset contains electronic health record (EHR) information from over 110,000 patients with urinary tract infections (UTI) treated at Massachusetts General Hospital and Brigham \& Women's Hospital in Boston, MA, USA between 2007 and 2016.
Each patient in the dataset provided urine cultures for antibiotic drug resistance testing.

We include only the observations with empiric antibiotic prescriptions from the AMR-URI dataset.
Exactly one of the first-line antibiotics, nitrofurantoin (NIT) or TMP-SMX (SXT), or one of the second-line antibiotics, ciprofloxacin (CIP) or levofloxacin (LVX) was prescribed.
Here, 11136 observations composed our dataset.
We remove observations that do not have any health event.
For each observation, a feature is constructed from its EHR as a binary indicator for whether the patient was undergoing a particular medical event in a specified time window.
The first part of the medical event is the past clinical history associated with antibiotic resistance, including recurrent UTIs, hospitalizations and resistance of previous infections. 
Besides, the risk of an infection being resistant to different antibiotics is associated with patient demographics and comorbidities. 
Known comorbidities associated with resistance include the presence of a urinary catheter, immunodeficiency and diabetes \cite{ikram2015outbreak}. 
Surgery, placement of a central venous catheter (CVC), mechanical ventilation, hemodialysis, and parenteral nutrition were included in the prior procedure description.



\subsection{Experiments Setting}
We split the dataset into train/validation/test subsets with 0.7/0.1/0.2 proportion. 
When showing experiment results, we compare the proposed method against logistic regression (LogReg), random forest(RF), support vector machine (SVM) and multi-layer perceptron (MLP) as single task model baselines. 
We also conduct extensive experiments to evaluate the combination of different encoders (i.e., GIN, GCN, GraphSAGE, SGC, GAT), multi-task optimization strategies (i.e., Dynamic Weight Adaptioin and Fixed Uniform Weight), and virtual node (yes/no). 
We set the number of convolution layers of GIN to 7, 
the neighbor pooling and readout function to \emph{sum pooling} and \emph{mean readout}, respectively. 
We use the configurations from their respective publications and set the number of layers of GCN, GAT, and SGC to 2 and GraphSAGE to 3. 
Batch size is set to 32 and learning rate to 0.01 after trial-and-error tuning. 
The strategy of learning rate decaying is step decaying.

\begin{figure}[t]
\centering
\includegraphics[width=1.0\linewidth]{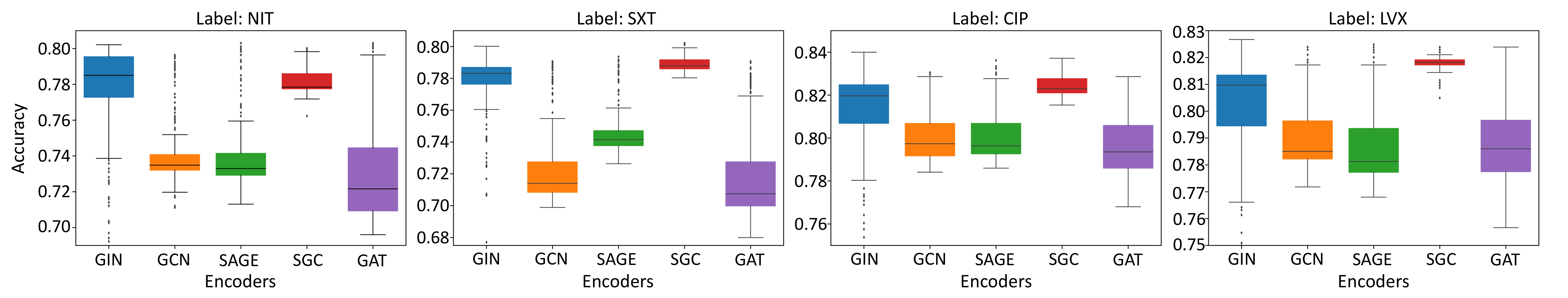}
\caption{
Comparison between the proposal (GIN) and four other graph encoders.
}
\label{fig:encoder_ablation}
\end{figure}

\begin{table}[t]
\centering
\caption{Comparison between the proposed method against conventional machine learning methods.
Here, $\uparrow$ refers to an improvement between the optimal convention and ours, $\downarrow$ refers a lower performance, while $\updownarrow$ denotes a competitive result.
}
\label{tb:comparison}
\resizebox{\linewidth}{!}{%
\begin{tabular}{llllllllll}
\toprule
\multirow{2}{*}{Methods/Labels~} & \multicolumn{2}{l}{~~~~~~NIT} & \multicolumn{2}{l}{~~~~~~SXT} & \multicolumn{2}{l}{~~~~~~CIP} & \multicolumn{2}{l}{~~~~~~LVX} \\
\cline{2-9}
& Recall & F1 & Recall & F1 & Recall & F1 & Recall & F1 \\
\midrule
LogReg & 0.16 & 0.47 & 0.19 & 0.47 & 0.39 & 0.61 & 0.44 & 0.63 \\
RF & 0.29 & 0.48 & \textbf{0.29} & \textbf{0.49} & 0.39 & 0.57 & 0.46 & 0.61 \\
SVM & 0.01 & 0.40 & 0.08 & 0.42 & 0.36 & 0.60 & 0.38 & 0.61 \\
MLP &  0.10 & 0.45 & 0.21 & 0.44 & 0.37 & 0.60 & 0.44 & 0.63 \\
Our Method & \textbf{0.65} (0.36)$\uparrow$ & \textbf{0.60} (0.12)$\uparrow$ & 0.20 (0.09)$\downarrow$ & 0.48 (0.01)$\updownarrow$ & \textbf{0.59} (0.20)$\uparrow$ & \textbf{0.69} (0.08)$\uparrow$ & \textbf{0.57} (0.11)$\uparrow$ & \textbf{0.68} (0.05)$\uparrow$ \\
\bottomrule
\end{tabular}
}
\end{table}

\subsection{Results and Analyses}
We firstly compare against other conventional ML approaches in Table \ref{tb:comparison}.
As expected, the baselines performed poorly in obtaining a sensible trade-off between precision and recall due to the high dimensionality and sparsity of the data. Our proposed method significantly outperformed the conventional methods on three label systems (NIT, CIP, and LVX), which is 0.36, 0.20, and 0.11 within Recall superior to the experimental conventional methods.
In contrary, on the SXT system the proposed method performed only slightly under RF: 0.09 lower in Recall and 0.01 lower in F1 score.

The characteristics of the data pose a challenge to graph encoders. To thoroughly investigate which encoder performs the best, we evaluate 5 popular GNN encoders on the validation set in Figure \ref{fig:encoder_ablation}. 
The result shows that GIN outperformed all other encoders on NIT label system and was competitive on the other label systems. 
Therefore, we might safely put that GIN is one of the most suitable graph encoders for the chosen scenario. 

\begin{figure}[t]
\centering
\includegraphics[width=1.0\linewidth]{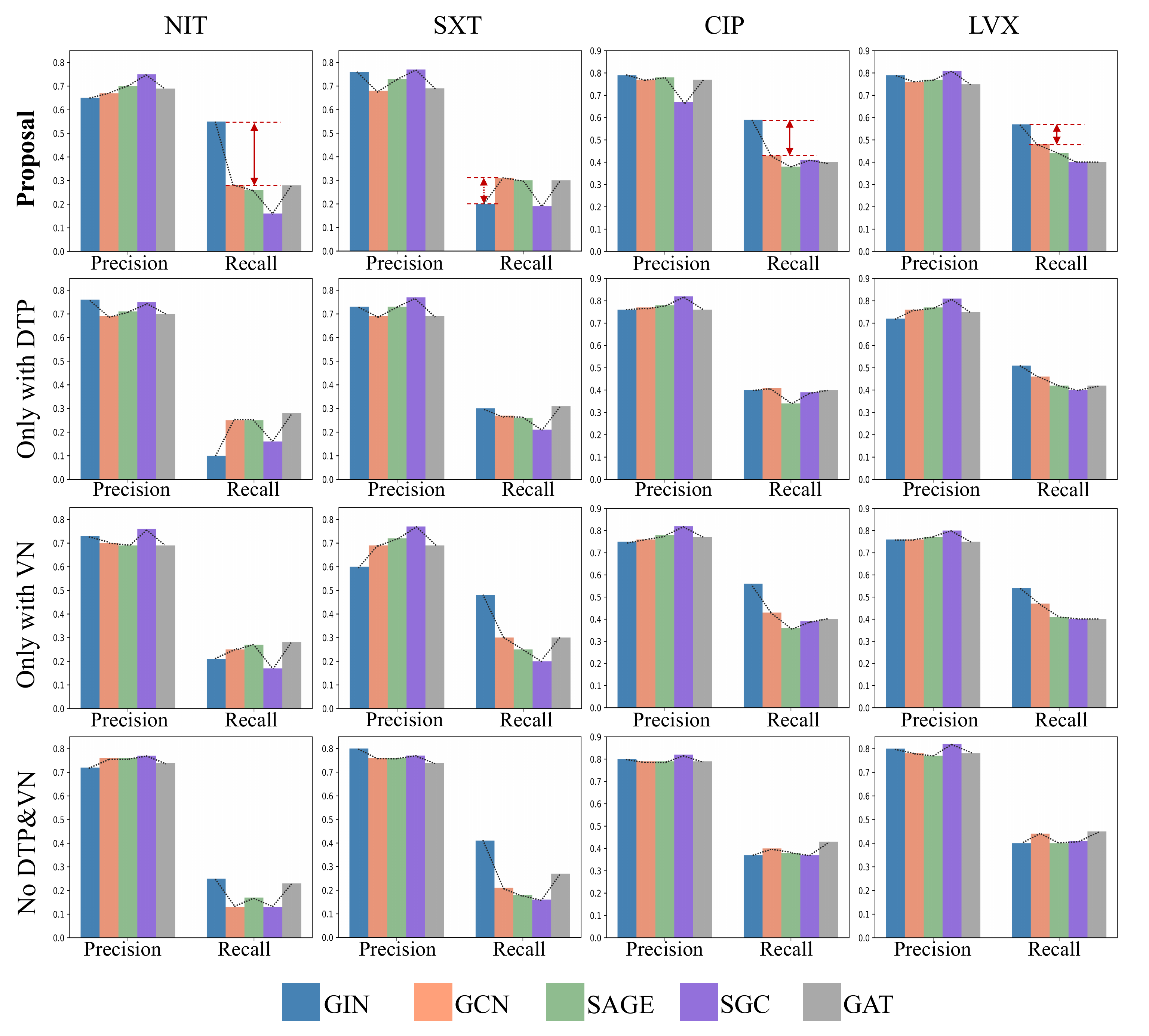}
\caption{
Comparison between all possible combinations of encoders, optimization strategy and the virtual nodes.
}
\label{fig:ablation}
\end{figure}

Figure \ref{fig:ablation} displays the result of our comprehensive ablation study.
We showed \emph{precision} and \emph{recall} of the four tasks on the testing set leveraging all possible combinations of graph-based encoders, multi-objective optimization strategies, and virtual node. 
As shown in the first column of Figure \ref{fig:encoder_ablation}, our proposed method significantly improved the Recall on NIT, CIP, and LVX. At the same time, it retained as much precision as possible at the cost of only slight loss on SXT Recall, which implies our proposed method can effectively alleviate the problem of label imbalance to exert the few positive samples.
Compared with the ablation choices, it is visible that 
 their performance suffered from significant drop. 
 This observation suggests the importance of DTP and VN without which the performance suffered.
 Hence it can be concluded that DTP and VN improve the proposed method from both the optimization-level (i.e., DTP dynamic control the learning process) and the model-level (i.e., virtual node play as a global agent to connect the isolate component).


\begin{figure*}[t]
\centering
\includegraphics[width=1.0\linewidth]{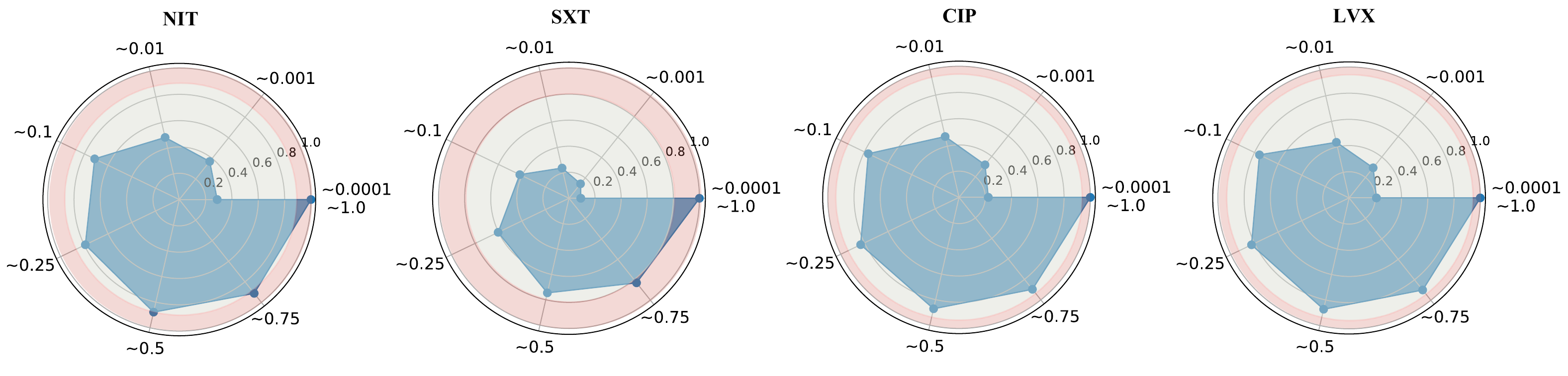}
\caption{Cumulative probabilities of drug resistance output by the model. 
Numbers around the circles indicate thresholds.
The two-color background represents the proportion of non-susceptible phenotype decided by antibiotic drug resistance testing in microbiology laboratory.
}
\label{fig:radar}
\end{figure*}

\subsection{Drug Resistance Phenotype Classification}

We visualize the drug resistance phenotypes in Fig. \ref{fig:radar} by thresholding probabilities output by the model. 
For different observation we set a series of distinct thresholds for probabilities.
Observations with model probabilities above the threshold is classified as \emph{non-susceptible} and \emph{susceptible} for those below the threshold. 
As the threshold increases, more observations are classified into the \emph{non-susceptible} phenotype.
The two-color background of the radar graphs represents the proportion of \emph{non-susceptible}  phenotype to the susceptible one, decided by antibiotic drug resistance testing in microbiology laboratory (NIT: 88.17\% SXT: 79.57\%, CIP: 93.88\%, LVX: 94.03\%).
A critical threshold allows the model phenotype distribution to approximate the laboratory phenotype distribution.
Such a threshold holds the promise to translate model probabilities into direct drug treatment recommendations.


\section{Conclusion}

In this paper we proposed to view health records consisting of binary or categorical medical events as graphs: nodes are mapped from the events and edges reflect whether two events happened at the same time.
A novel graph-based deep learning architecture was then proposed to extract informative features such as correspondence between medical events from those high dimensional and sparse graphs.
On a massive dataset comprising over 110,000 patients with urinary tract infections we verified that the proposed method was capable of attaining superior performance on the drug resistance prediction problem.
We further showed that automated drug recommendations could be made on top of the resistance analyses output by the model.
Extensive ablation studies within the graph neural networks literature as well as against conventional baselines demonstrated the effectiveness of the proposed binary health record data as graphs framework.

 
%
%
%
\bibliographystyle{splncs04}
\bibliography{Bibliography}
%




\end{document}